\documentclass[10pt,twocolumn,letterpaper]{article}

\usepackage{ijcb}
\usepackage{times}
\usepackage{epsfig}
\usepackage{graphicx}
\usepackage{amsmath}
\usepackage{amssymb}
\usepackage{bm}
\usepackage{subfig}
\usepackage{caption}
\usepackage{diagbox}
\usepackage{balance}
\usepackage[norule,symbol,perpage]{footmisc}
\graphicspath{{IJCBIMAGES/}}

\usepackage[pagebackref=true,breaklinks=true,letterpaper=true,colorlinks,bookmarks=false]{hyperref}

\ijcbfinalcopy % *** Uncomment this line for the final submission

\ifijcbfinal\fi

\makeatletter

\def\ps@IEEEtitlepagestyle{
\def\@oddfoot{\mycopyrightnotice}
\def\@evenfoot{}
}
\def\mycopyrightnotice{
{\hfill \footnotesize 978-1-6654-3780-6/21/\$31.00 \copyright 2021 IEEE\hfill}
}
\makeatother

\makeatletter

\newcommand*\titleheader[1]{\gdef\@titleheader{#1}}
\AtBeginDocument{%
  \let\st@red@title\@title
  \def\@title{%
    \bgroup\normalfont\normalsize\centering\vspace{-3cm}\@titleheader\par\egroup
    \vskip2em\st@red@title}
    }
\makeatother

\title{Conditional Identity Disentanglement for Differential Face Morph Detection}
\titleheader{\textcolor{red}{S. Banerjee and A. Ross, ``Conditional Identity Disentanglement for Differential Face Morph Detection," Accepted in International Joint Conference on Biometrics (IJCB), (Schenzhen, China), August 2021}}
\author{Sudipta Banerjee and Arun Ross\\
Michigan State University\\
{\tt\small banerj24@cse.msu.edu, rossarun@cse.msu.edu}}
\begin{document}
\maketitle
\thispagestyle{empty}

%%%%%%%%% ABSTRACT
\begin{abstract}
We present the task of differential face morph attack detection using a conditional generative network (cGAN). To determine whether a face image in an identification document, such as a passport, is morphed or not, we propose an algorithm that learns to implicitly disentangle identities from the morphed image conditioned on the trusted reference image using the cGAN. Furthermore, the proposed method can also recover some underlying information about the second subject used in generating the morph. We performed experiments on AMSL face morph, MorGAN, and EMorGAN datasets to demonstrate the effectiveness of the proposed method. We also conducted cross-dataset and cross-attack detection experiments. We obtained promising results of 3\% BPCER @ 10\% APCER on intra-dataset evaluation, which is comparable to existing methods; and 4.6\% BPCER @ 10\% APCER on cross-dataset evaluation, which outperforms state-of-the-art methods by at least 13.9\%. 
\end{abstract}

\renewcommand*{\thefootnote}{\arabic{footnote}}

%%%%%%%%% BODY TEXT
\section{Introduction}
\label{Sec:Intro}

\begin{figure}
    \centering
    \includegraphics[width=0.485\textwidth]{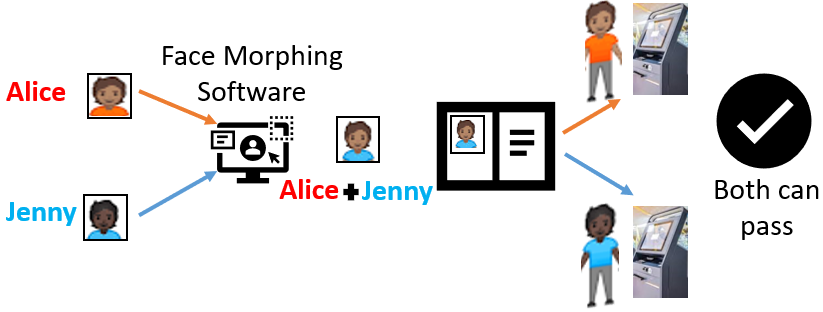}
    \caption{Illustration of face morphing exploited to provide access to two different subjects, Alice and Jenny. Both can use the same morphed image on the passport at an airport e-gate for access.}
    \label{fig:Morphing}
    
\end{figure}

\begin{figure}
    \centering
    \includegraphics[width=0.48\textwidth]{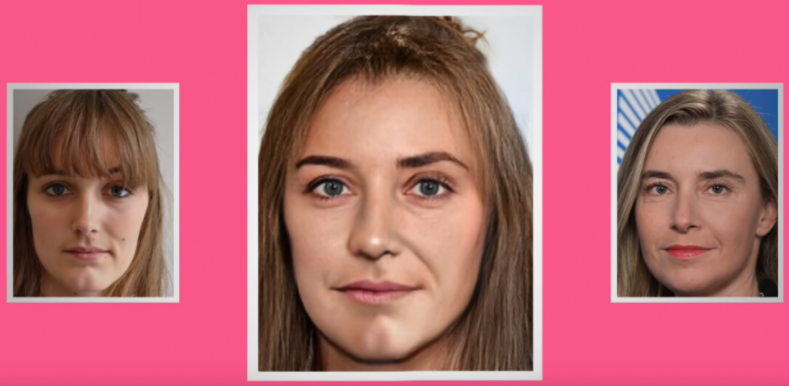}
    \caption{The morphed image in the middle that was used to apply for a passport in the EU was created using the photograph of an activist (left) and an official (right). Image courtesy: Peng!(MaskID) reproduced from~\cite{Germanmorphing}.}
    \label{fig:realworld}
   
\end{figure}

Face morphing involves a continuous transition from the face image of one individual (source identity) to the face image of the second individual (target identity)~\cite{History}. The idea of morphing has been used for visual effects in entertainment videos. But recently, it has been demonstrated that face morphing can be used for adversarial purposes. Since a morphed face contains features from two individuals, it can successfully match both identities, thereby posing a security threat. See Figure~\ref{fig:Morphing}. This problem is of practical concern due to two reasons: 1) the prolific use of biometric data in official documents for authentication in unattended border control systems (viz., face images in passports accepted at e-gates), and 2) the ease of access to face image editing software (e.g., FaceMorpher~\cite{FaceMorpher}). The idea of using morphed face image in identity documents was first postulated in~\cite{Magic}.\footnote{Note that face morphing can be used in a positive manner also in privacy-preserving applications~\cite{Asem}.} 
{\let\thefootnote\relax\footnotetext{\mycopyrightnotice}}
Later, a real-world case of face morphing was reported where an art activist morphed her face image with that of the photograph of an EU Foreign Affairs Commissioner to apply for a German passport, prompting the authorities to reconsider using self-created digital photographs in identity documents~\cite{Germanmorphing}. See Figure~\ref{fig:realworld}. Face morph attack has piqued the interest of the research community and government agencies alike, leading to the investigation of methods that can not only produce realistic face morphs but also successfully detect such types of morphs. The EU-funded iMARS project is geared towards developing image morphing techniques and manipulation attack detection solutions for identification documents~\cite{iMARS}. Note that current deep-fake detectors cannot effectively discriminate between morphed and legitimate bonafide (non-morphed) images. Further, existing anti-spoofing solutions developed for presentation attacks are often not suited to detect morph attacks since the latter are {\em digital} alterations rather than {\em physical} presentations.  
%Even the best performing open-source face matcher, such as ArcFace, fails to distinguish between genuine and morphed images. 

Face morph generation typically involves two kinds of approaches: 1) landmark-based approaches, and 2) generative model-based approaches. The first approach utilizes facial landmarks from two face images for aligning them using warping~\cite{LMA1, LMA2}. Image blending principles are then used to combine the pixels in the overlapped regions to construct the morphed image. An optional post-processing step involving histogram equalization or Poisson blending may be applied to achieve visual realism and to remove ghosting artefacts. The second approach leverages the generative capability of adversarial networks to synthesize morphs. MorGAN~\cite{MORGAN}, MIPGAN~\cite{MIP} and EMorGAN~\cite{CIEMORGAN} are examples of such approaches. The final morphed image can then be inserted into an official document used at access control points. The morphs can be so realistic that visual inspection alone may not be sufficient for detecting them. See Figure~\ref{fig:realworld}. 

Face morph attack detection (MAD) can be performed using two methods: 1) reference-free single image-based MAD, and 2) reference-based differential MAD. The former tries to address the following question. \textit{Given a document face image can we determine whether it is morphed or not?} According to the NIST FRVT report~\cite{NIST}, the best performing algorithm in this category results in 93.8\% APCER @ 10\% BPCER on high quality morphs in ``Dataset:Manual"~\cite{NIST}. Here, bonafide presentation classification error rate (BPCER) denotes the proportion of bonafide (non-morphed) images incorrectly classified as morphed images, and attack presentation classification error rate (APCER) denotes the proportion of morphed images incorrectly classified as non-morphed images. The second method tries to address the following question. \textit{Given a document face image, and a live capture as a reference image, can we determine whether the document image is morphed or not?} According to the NIST FRVT report~\cite{NIST}, the best performing algorithm in this category results in 9.1\% APCER @ 10\% BPCER on high quality morphs in ``Dataset:Manual". It is evident, based on these results, that the state-of-the-art performance requires significant improvement. The readers are referred to surveys in~\cite{Survey1, Survey2} for a comprehensive overview of face morph generation and detection. In this paper, we focus on developing a novel \textit{differential MAD method} to advance the state of the art. \textbf{The novelty of the method lies in formulating the differential MAD problem using an information theoretic framework for a sound detection strategy.}

The remainder of the paper is organized as follows. Section~\ref{Sec:Rel} describes existing differential MAD strategies. Section~\ref{Sec:Prop} describes the proposed method. Section~\ref{Sec:Expts} outlines the experiments. Section~\ref{Sec:Results} reports and analyzes the results. Finally, section~\ref{Sec:Sum} concludes the paper with summary and future directions.

\section{Related Work}
\label{Sec:Rel}
Differential MAD strategies use a reference-based approach. Here, the trusted live face image of a subject taken during the time of acquisition (called the reference image) is used along with the document face image (in the passport, for example) to determine whether the latter is morphed or non-morphed. The first known work on differential MAD performs ``demorphing'' which uses the difference computed between the document image and the reference image to ascertain whether the document image is morphed or not, and alerts the officer for additional inspection, if necessary~\cite{DMAD3}. Demorphing also tries to uncover the identity of the second subject. Later, GAN-based de-morphing has been proposed in the literature~\cite{FDGAN}. Another class of techniques utilizes features such as BSIF or the disparity between landmarks, along with a classifier, to detect morphs differentially~\cite{DMAD_BSIF,DMAD1,DMAD2}. The detection performance is further enhanced by the use of deep face representations~\cite{DMAD4}. Recently, a method that uses appearance and landmark disentanglement modules for differential MAD was proposed in~\cite{DMAD6}. It uses the disentanglement module to learn complementary information, and contrastive loss to boost detection performance. An approach referred to as Focused Layer-Wise Relevance Propagation (FLRP) aims at explaining the decision made by the deep neural network in detecting morphs for better interpretability~\cite{FocusedLRP}.

\section{Proposed Method}
\label{Sec:Prop}

\begin{figure}
    \centering
    \includegraphics[width=0.49\textwidth]{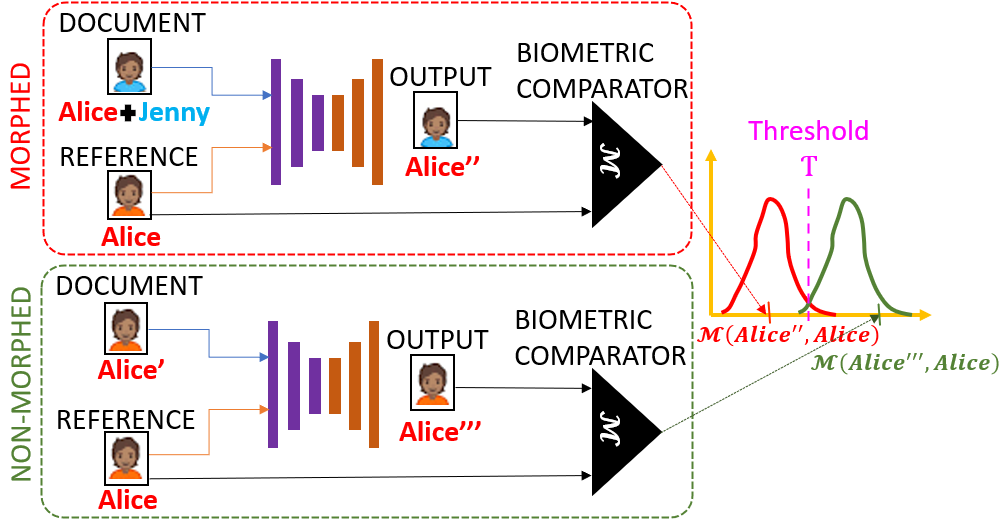}
    \caption{The proposed method uses a conditional disentanglement framework to discriminate between morphed and non-morphed document images conditioned on the reference image. The biometric comparator scores can be compared against a suitable threshold, $\mathcal{T}$, for differential morph attack detection.}
    \label{fig:Prop}
   
\end{figure}
\subsection{Conditional GAN}
The proposed method requires a tool that can convert a source image, $\bm{X}$, which is the `document' image, to a target image, $\bm{Y}$, which is the `reference' image. This can be achieved via image translation guided by a conditional generative adversarial network (cGAN)~\cite{cgAN}. It uses the following objective function.
\begin{align}
    G^* &= \displaystyle\arg\displaystyle\min_G\displaystyle\max_D \{ \mathbb{E}_{(\bm{X},\bm{Y})}[\log D(\bm{X},\bm{Y})] \nonumber \\ &+  \mathbb{E}_{(\bm{X},\bm{z})}[\log(1-D(\bm{X},G(\bm{X},\bm{z})))] \} \nonumber \\ &+ \lambda \mathbb{E}_{(\bm{X},\bm{Y},\bm{z})}[\|(\bm{Y} - G(\bm{X},\bm{z}))\|_1]
\end{align}

% \begin{align}
%     \displaystyle\min_G\displaystyle\max_D V(D,G) &= \mathbb{E}_{x\sim p_{data}(x)}[\log D(x \mid y)] \\ &+  \mathbb{E}_{z\sim p_{z}(z)}[1-\log(D(G(z \mid y)))]
% \end{align}

Here, $D$ refers to the discriminator and $G$ refers to the generator. The two inputs to the network are $\bm{X}$ (document image) and $\bm{Y}$ (reference image). The reference image is always assumed to be a bonafide since it is captured ``live''. Now the network has to learn to translate from $\bm{X}$ to $\bm{Y}$. The Gaussian random noise vector $\bm{z}$ regularizes the training and diversifies the output. If $\bm{X}$ is non-morphed, then the task of the generator is to simply reconstruct itself with some additional variations due to age, pose, illumination and expression, which can be deduced from $\bm{Y}$. However, if $\bm{X}$ is morphed, the challenge is two-fold: (i) remove traces of the second identity, and (ii) incorporate variations related to pose, age, expression and appearance. We train the network with both types of examples: $(\bm{X}_{\tiny{non-morphed}}, \bm{Y})$ and $(\bm{X}_{\tiny{morphed}}, \bm{Y})$, i.e., (non-morphed source and bonafide target images) and (morphed source and bonafide target images). \textbf{However, the method does not require any labels (supervision) regarding which pairs correspond to morphed source images and which pairs correspond to non-morphed source images.} The idea is to make the network automatically learn to disentangle the composite (morphed) image conditioned on the reference image. If the image is not a composite, the task is simpler, it has to generate a variant of the source image.
\subsection{Rationale of the Proposed Method}
\label{Rationale}
The majority of the literature poses differential MAD as a supervised classification problem, with the underlying premise that discriminative features can be deduced using either pre-defined filters (hand-crafted) or automatically learned filters (deep-learning). In contrast, we make the case that the task of differential MAD can be framed in an information theoretic framework that describes the translation from the document image to the reference image. 

\begin{align}
    \mathcal{H}(\bm{X} \mid \bm{Y}) &= \mathcal{H}(\bm{Y}) - \mathcal{H}(\bm{X},\bm{Y})
\end{align}

If $\bm{X}$ is non-morphed, then $\bm{Y} = [\bm{X}+\delta]$, where $\delta$ captures the intra-class variations between document and reference images belonging to the same subject. Therefore, the conditional entropy can be formulated as,
\begin{align}
    \mathcal{H}(\bm{X} \mid [\bm{X}+ \delta])= \mathcal{H}([\bm{X} + \delta]) -  \mathcal{H}(\bm{X},[\bm{X}+ \delta]) 
\end{align}
Clearly the uncertainty associated with inferring $\bm{X}$ from a slightly noisy or altered version of itself ($\delta$ can be interpreted as additive noise) will be low. On the contrary, for a morphed image comprising two statistically independent distinct subjects, the uncertainty will be relatively higher due to the presence of a second identity. 
\begin{align}
\label{Eqn10}
    \mathcal{H}([\bm{X}_1, \bm{X}_2] \mid [\bm{X}_1+ \delta]) &= \nonumber \\ \mathcal{H}([\bm{X}_1 + \delta]) &-  \mathcal{H}(\bm{[\bm{X}_1, \bm{X}_2]},[\bm{X}_1+ \delta]) 
\end{align}
In Eqn.~(\ref{Eqn10}), we assume subject $\bm{X}_1$ appears at the verification checkpoint, so the uncertainty arises due to $\bm{X}_2$. A similar situation arises when the roles are reversed. 
Therefore, we expect $\displaystyle \mathcal{H}(\bm{X}_{morphed} \mid \bm{Y})>\mathcal{H}(\bm{X}_{non-morphed} \mid \bm{Y})$. Here, we interpret entropy loosely as the disparity between the output of the cGAN and the reference image.
Next, we describe the two steps in which we implement the proposed method.

In the first step, we use the cGAN to translate the source image (document) to the target image (reference): $\bm{X} \rightarrow \bm{Y}$. If $\bm{X}$ is non-morphed and represents the same individual as $\bm{Y}$, then the output (translated) image will be \textit{more similar} to $\bm{Y}$. On the other hand, if $\bm{X}$ is morphed, implying that it comprises two identities ($\bm{X}_1,\bm{X}_2$), then the output image will be \textit{less similar} to $\bm{Y}$. The translation will force the network to implicitly learn to disentangle identities. This is because, in order to translate from the source image (two identities if morphed) to the target image (only one identity), i.e., $[\bm{X}_1,\bm{X}_2] \rightarrow \bm{X}_2$ (or $\bm{X}_1$), it will try to remove traces pertaining to the second identity not present in the reference image, i.e., $\bm{X}_1$ (or $\bm{X}_2$), thereby striving to decouple the two identities present in the morphed image. We refer to this as \textit{conditional identity disentanglement}, where we disentangle the identities from a morphed image conditioned on a reference image. A fortuitous outcome of this process is that we can use the same method for deciphering information about the second subject from the morphed image. This is the novelty of the proposed method. By positing the differential morph detection problem using an information theoretic framework, not only can we detect morphs, but also disentangle the identities.
But how do we quantify the disparities between the output of the cGAN and the reference image to deduce whether a document image is morphed or not? 

In the second step, we use the output, $\bm{O}=G(\bm{X},\bm{z})$, of the cGAN, where $\bm{z}$ is the random noise vector, and compare it with the reference image, $\bm{Y}$, using a biometric comparator, $\mathcal{M}$. The score produced after the comparison can then be compared against a user-defined threshold, $\mathcal{T}$, that regulates error rates for the intended application to make the final decision: 
\begin{align}
 \hfill &\mathcal{M}(G(\bm{X},\bm{z}), \bm{Y}) 
    = \mathcal{M}(\bm{O}, \bm{Y}); \\
\noindent \bm{X} &= 
\begin{cases} 
\mbox{Morphed,} & \mbox{if}  \hspace{0.2cm}\mathcal{M}(\bm{O}, \bm{Y}) <\mathcal{T}, \\
\mbox{Non-morphed,} & \mbox{otherwise.}
\end{cases}
\end{align}
Figure~\ref{fig:Prop} outlines the proposed method.

\section{Experiments}
\label{Sec:Expts}

We describe the (i) datasets, (ii) conditional GAN setup, and (iii) experimental protocols used in this work.
\subsection{Dataset}
\label{SubSec:Data}

We used three datasets in this work. (i) \textbf{AMSL face morph dataset}~\cite{AMSL1, AMSL2}: It contains images from 102 subjects captured with neutral as well as smiling expressions. There are 2,175 morphed images corresponding to 92 subjects created using a landmark-based approach. In our problem formulation, the document image can be either (a) an image with neutral expression which will be considered as the bonafide, or (b) a morphed image. The reference image is a trusted live capture (bonafide) corresponding to the image of the same subject but with a smiling expression. (ii) \textbf{MorGAN dataset}~\cite{MORGAN}: It contains 500 bonafide images. Two morphs are generated using a generative network for each of these bonafide images from the two subjects most similar to the bonafide image resulting in 1,000 morphed images, which were split into train and test sets. (iii) \textbf{EMorGAN dataset}~\cite{CIEMORGAN}: It uses a cascaded image enhancement network to improve the quality of the morphed images synthesized using MorGAN, and has the same train and test split as~\cite{MORGAN}. 
\subsection{Implementation Details}
\label{SubSec:Imp} 

We explored different options for conditional GANs (cGANs) in the literature and found PIX2PIX~\cite{PIX2PIX} to be a suitable choice for this work. PIX2PIX uses a cGAN to translate images from one domain (e.g., sketch) to another domain (e.g., photorealistic images). PIX2PIX~\cite{PIX2PIX} follows $Conv \rightarrow BatchNorm \rightarrow ReLU$ architecture both in the generator and discriminator. The discriminator loss function minimizes the difference between the real and the fake images. The generator loss function maximizes the log-likelihood of the generated images while ensuring they are as close as possible to the target images using $\mathcal{L}_1$ loss. Readers are referred to~\cite{PIX2PIX} for additional details about the implementation. We used mini-batch stochastic gradient descent optimization algorithm with Adam solver at an initial learning rate of $\displaystyle 2\times 10^{-4}$ and momentum parameter of $\displaystyle 0.5$, and trained for 50/100/200 epochs (compared to 600K iterations in ~\cite{DMAD6}). 

\begin{figure*}
    \centering
    \includegraphics[width=0.85\textwidth]{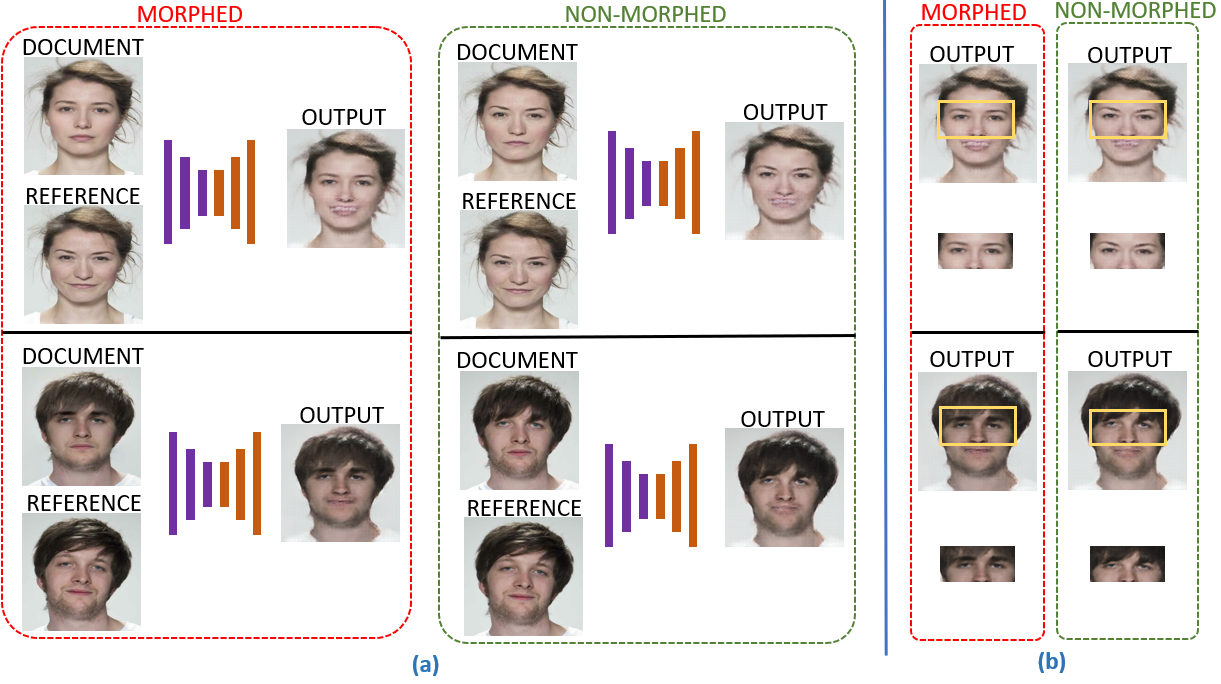}
    \caption{(a) Illustration of conditionally disentangled outputs generated using the proposed method for two subjects (top and bottom) belonging to the AMSL dataset when their document images were morphed (red) and non-morphed (green). (b) Zoomed-in view of the periocular regions of the generated outputs to emphasize the difference between morphed and non-morphed images.}
    \label{fig:Expt1}
   
\end{figure*}

\subsection{Experiments}
\label{SubSec:Proto} 
We conducted three experiments designed to answer the following research questions. 

\textbf{Experiment 1:} \textit{How does the proposed differential MAD method perform?} \\
To answer the above question, we trained on the AMSL dataset using 120 pairs of document and reference images out of which 60 pairs were morphed document images and the other 60 pairs were bonafide document images. These 60 subjects constitute 65\% of the 92 subjects, while the test set comprised of 778 images (745 morphed and 33 bonafide images) corresponding to the remaining 35\% of the subjects. \textbf{During training, no label is required to indicate which pairs correspond to morphed images and which pairs correspond to non-morphed images.} We feed the images generated by our network and the target reference images to a face recognition system (a COTS face comparator) and use the scores to determine whether a document image is morphed or not. We selected a subset of training images provided by the authors in the MorGAN~\cite{MORGAN} and the EMorGAN~\cite{CIEMORGAN} datasets. We trained on 120 pairs (60 pairs of bonafide document images and 60 pairs of morphed document images), and used the entire test set for evaluation in both cases.

\textbf{Experiment 2:} \textit{Is the proposed method generalizable under cross-dataset and cross-attack scenarios?} \\
To answer the above question, we trained on one dataset (one type of attack) and tested on the remaining two datasets (other types of attacks). AMSL dataset represents a landmark-based morph attack while MorGAN and EMorGAN datasets represent generative model-based attacks.

\textbf{Experiment 3:} \textit{Can the method be used for recovering some discriminative information about the second subject?}\\
To answer the above question, we used a different training strategy compared to morph detection. In this experiment, we used the pixel-wise difference image computed between the document and the reference image from the AMSL dataset as the source image, and the document image as the target image. The intuition behind this strategy is to ensure that the network learns to map the residual (reminiscent of the second subject contributing to the morph) to the morphed target image.

\section{Results and Analysis}
\label{Sec:Results}
We report both qualitative and quantitative results of the proposed method. We report the results in terms of the metrics predominantly used in the morph detection literature: BPCER @ APCER of 10\% following~\cite{DMAD6} as well as APCER @ BPCER of 10\%  following~\cite{NIST}.  
\begin{figure}[h]
    \centering
    \subfloat[Bonafide and Morphed-Before]
    {\centering
    \includegraphics[scale=0.35]{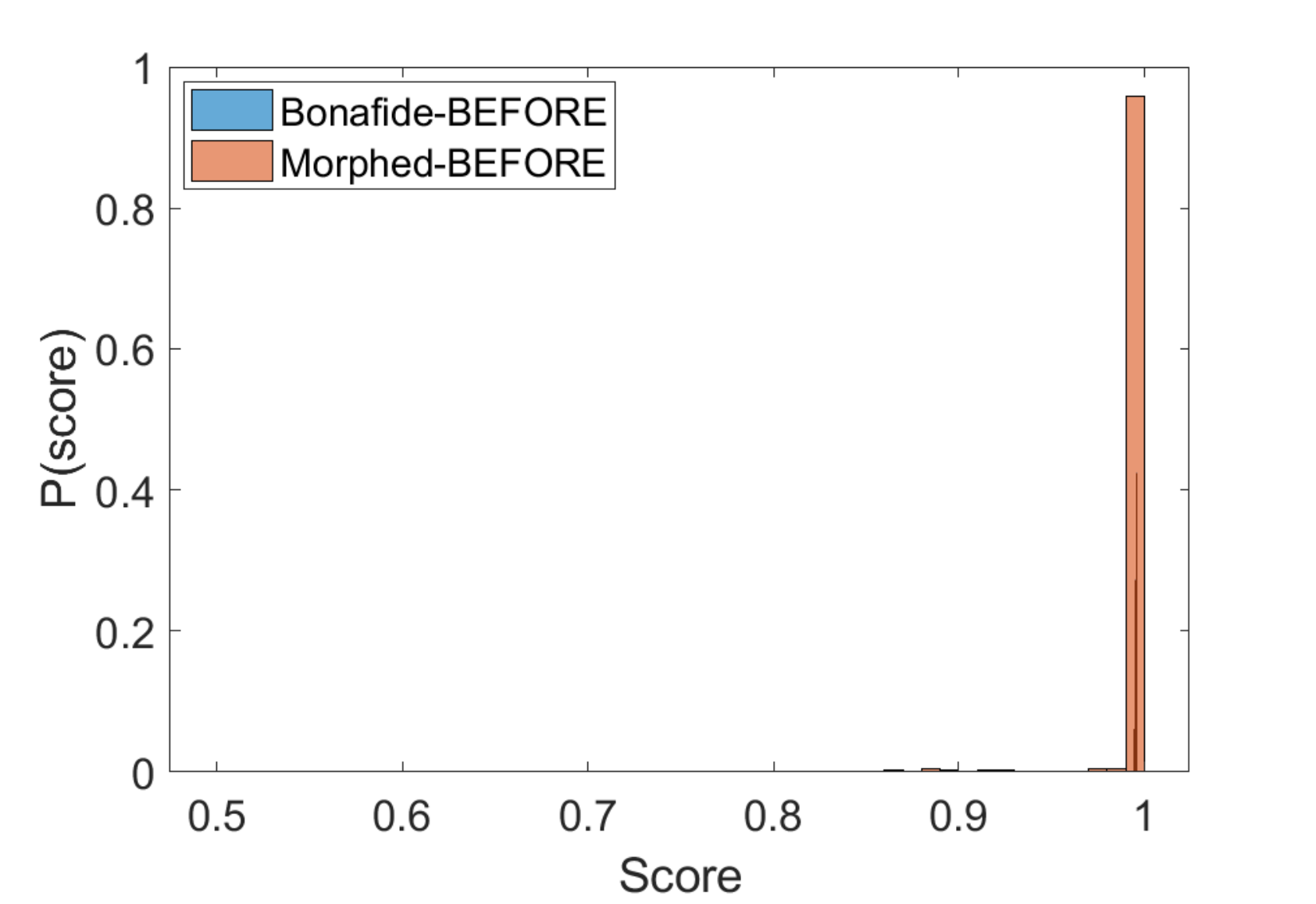}
    }\hspace{-.1cm}
    \subfloat[Bonafide and Morphed-After]
    {
    \centering
    \includegraphics[scale=0.35]{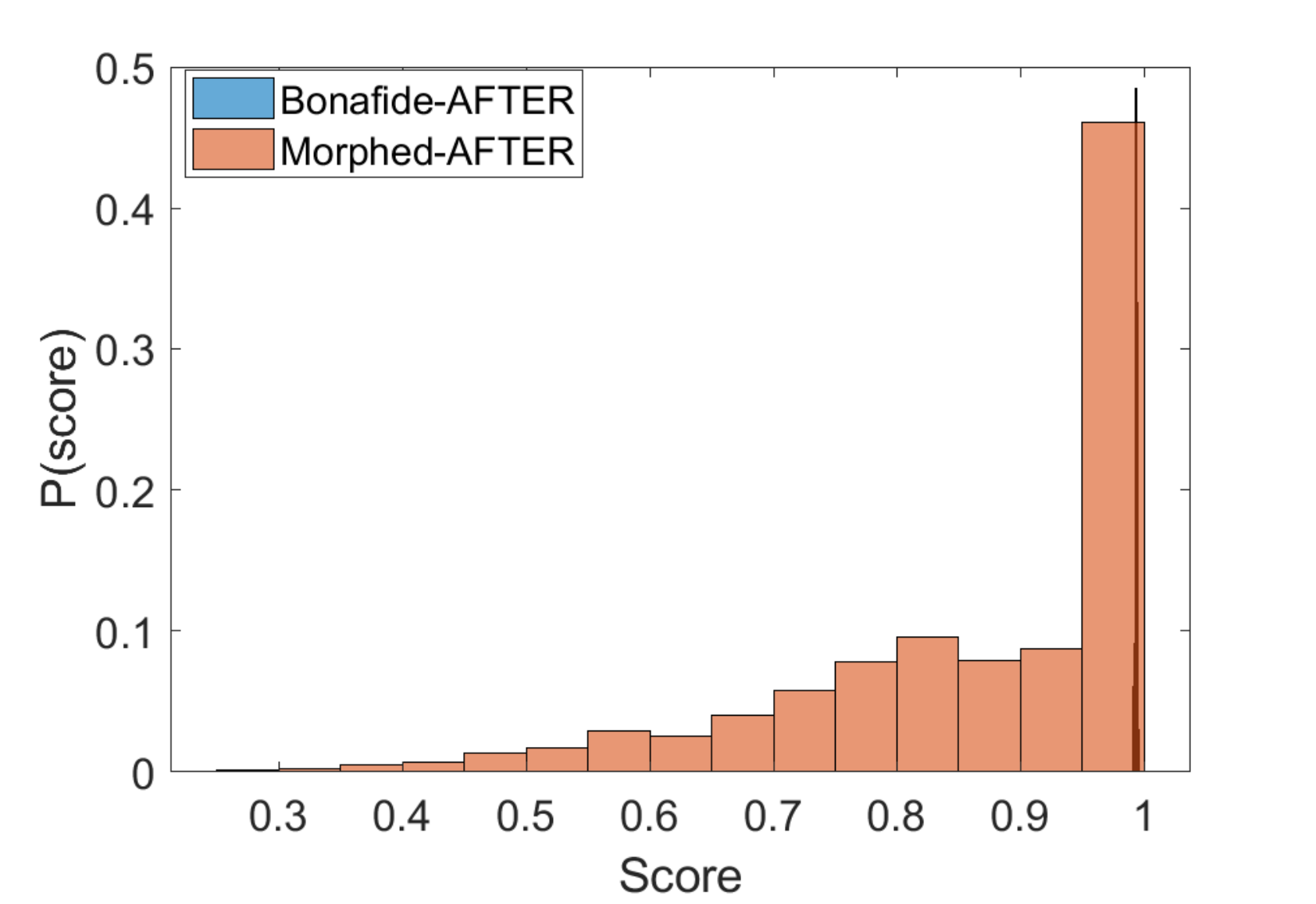}
    }\\
    \subfloat[Bonafide]
    {
    \centering
    \includegraphics[scale=0.35]{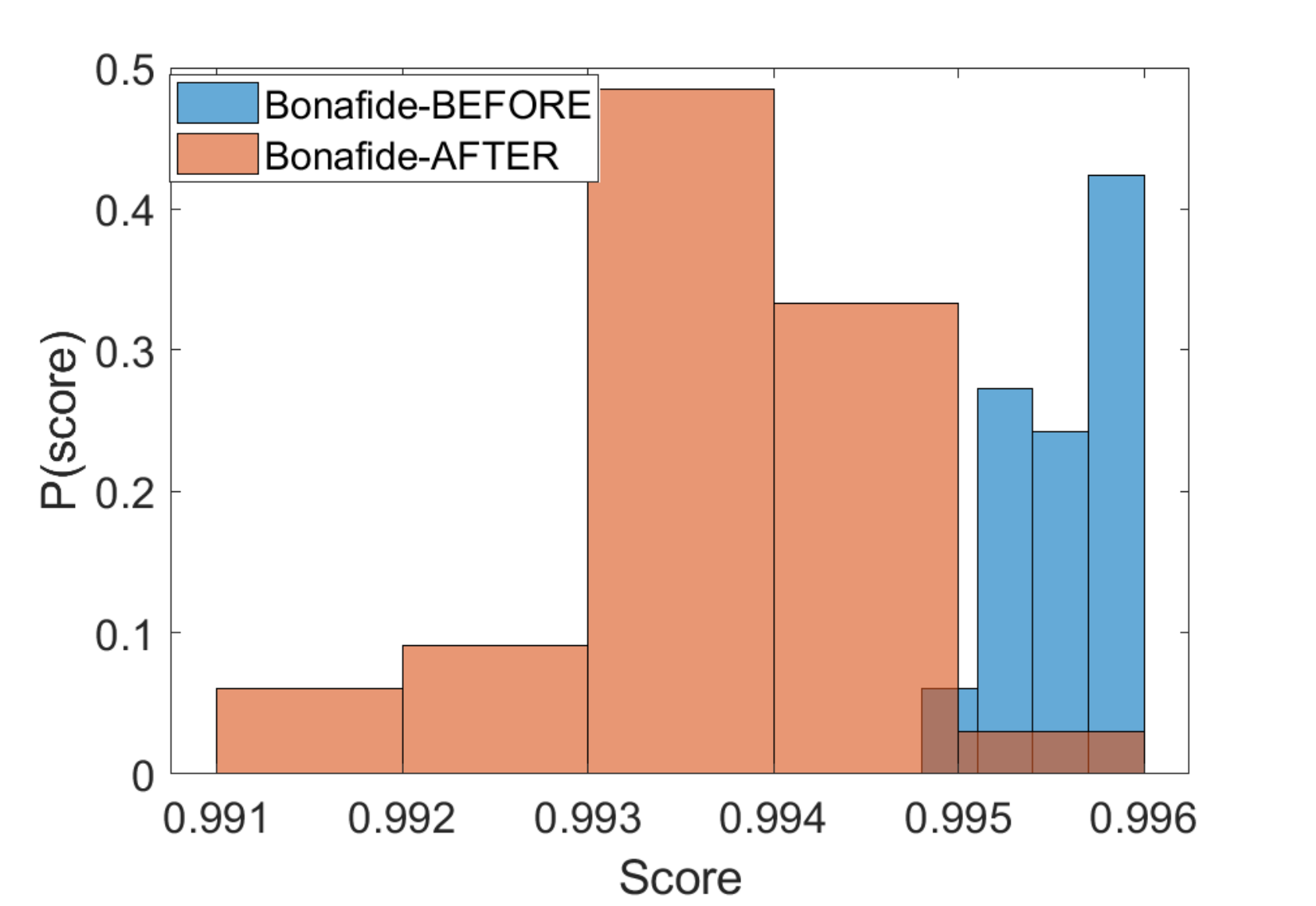}
    }\hspace{-.1cm}
    \subfloat[Morphed]
    {
    \centering
    \includegraphics[scale=0.35]{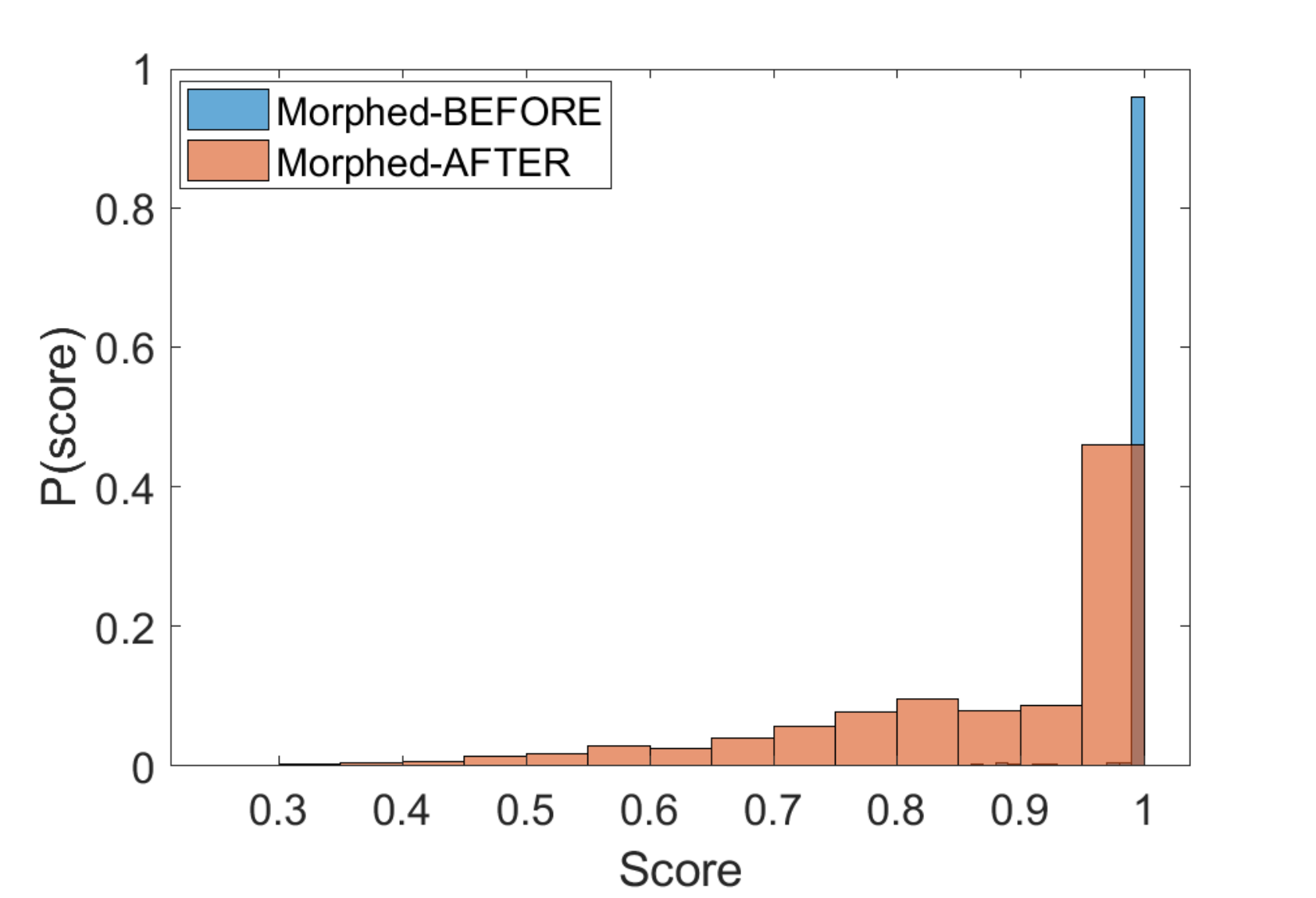}
    }
    \caption{Variations in score distributions before and after conditional disentanglement in the AMSL dataset indicating successful disentanglement in morphed document images (Experiment 1).}
    \label{fig:Histograms}
   
\end{figure}
  
  \textbf{Results from Experiment 1:} Table~\ref{Tab:ten} reports the performance of the proposed method. We would like to re-iterate that the proposed method was not trained to discriminate between non-morphed and morphed images, i.e., during training the method does not require labels corresponding to morphed and non-morphed image pairs. In spite of that, the proposed method achieves comparable results to some extent with state-of-the-art methods that require explicit supervision (see Table~\ref{Tab:comparison}). Qualitatively, we present the results in Figure~\ref{fig:Expt1}. In order to demonstrate that the proposed method is actually disentangling subject identities, we present the score distributions produced by the COTS comparator, when comparing test images belonging to the AMSL dataset (see Figure~\ref{fig:Histograms}). In Figure~\ref{fig:Histograms}(a), the distributions corresponding to morphed and bonafide images completely overlap (BEFORE disentanglement). In Figure~\ref{fig:Histograms}(b), the distribution corresponding to morphed images move towards zero (AFTER disentanglement). Here, the scores are similarity scores normalized to $[0, 1]$. To illustrate how disentanglement has a pronounced effect on morphed images compared to bonafide images, we present the score distributions of the \textit{bonafide} images before and after disentanglement in Figure~\ref{fig:Histograms}(c), and that of the \textit{morphed} images before and after disentanglement in Figure~\ref{fig:Histograms}(d). The bonafide scores are closely located (note the range of the scores in the x-axis in Figure~\ref{fig:Histograms}(c)). In contrast, if the document image is morphed, the network tries to disentangle the identity traces belonging to the second subject which causes a significant shift in the scores (see Figure~\ref{fig:Histograms}(d)). We further computed the interquartile range (IQR) as a `measure of dispersion' between the bonafide and morphed score distributions. This was done for bonafide and morphed scores both before and after disentanglement. Then we computed the ratio of their IQRs and observed that the ratio for morphed scores was $\sim$6 times higher than that of the bonafide scores. $\bigg(\frac{IQR_{morphed-AFTER}}{IQR_{morphed-BEFORE}} > 5.9\times \frac{IQR_{bonafide-AFTER}}{IQR_{bonafide-BEFORE}}\bigg)$.
  
  We observed from the experiments that the AMSL dataset required the least number of epochs followed by MorGAN and EMorGAN. In the case of AMSL, the BPCER improved from 6.1\% to 3\% upon increasing the number of training epochs from 50 to 200. But the BPCER remained the same when tested on the MorGAN and EMorGAN datasets.

% \begin{table}[]
% \centering
%  \caption{BPCER(\%) @ APCER=5\% (left to the forward slash) and APCER(\%) @ BPCER=5\% (right to the forward slash).}
%  \scalebox{0.95}{
% \begin{tabular}{|l||*{5}{c|}}\hline
% \backslashbox{Train}{Test}
% &\makebox[3em]{AMSL}&\makebox[3em]{MorGAN}&\makebox[3em]{\begin{tabular}[c]{@{}l@{}}CIE-\\ MorGAN\end{tabular}}
% \\\hline\hline
% AMSL &  &  &  \\\hline
% MorGAN &69.7/66.4 & &   \\\hline
% \begin{tabular}[c]{@{}l@{}}CIE-\\ MorGAN\end{tabular} &38.7/51.5 & & \\ \hline

% \end{tabular}}
% \label{Tab:one}
%  \end{table}

\begin{table}[]
\centering
 \caption{BPCER(\%) @ APCER=10\% (left of the forward slash) and APCER(\%) @ BPCER=10\% (right of the forward slash).}
 \scalebox{0.9}{
\begin{tabular}{|l||*{5}{c|}}\hline
\backslashbox{Train}{Test}
&\makebox[3em]{AMSL}&\makebox[3em]{MorGAN}&\makebox[3em]{\begin{tabular}[c]{@{}l@{}}EMorGAN\end{tabular}}
\\\hline\hline
AMSL (50 epochs) & 6.1/5.2 & 4.6/2.0 & 4.4/0.8 \\\hline
MorGAN (100 epochs) &63.6/60.3 & 8.6/5.8& 9.4/9.3  \\\hline
\begin{tabular}[c]{@{}l@{}}EMorGAN\end{tabular} (200 epochs) &25.8/47.7 &28.1/38.4 & 29.7/39.4  \\ \hline

\end{tabular}}
\label{Tab:ten}
 \end{table}

\begin{table}[]
\centering
 \caption{BPCER(\%) @ APCER=10\%. The best performing results for the proposed method are compared with the baseline results indicated within parentheses (\cite{DMAD6},~\cite{DMAD4}). The baseline results are taken from~\cite{DMAD6}.}
 \scalebox{0.95}{
\begin{tabular}{|l||*{4}{c|}}\hline
\backslashbox{Train}{Test}
&\makebox[3em]{AMSL}&\makebox[3em]{MorGAN}
\\\hline\hline
AMSL & 3.0 (\textbf{2.2}, 3.3)&\textbf{4.6} (18.5, 24.5)\\\hline
MorGAN &63.6 (\textbf{8.8}, 14.9) & \textbf{8.5} (\textbf{8.5}, 12.4)\\\hline

\end{tabular}}
\label{Tab:comparison}
 \end{table}
 
 \begin{figure}
    \centering
    \includegraphics[width=0.5\textwidth]{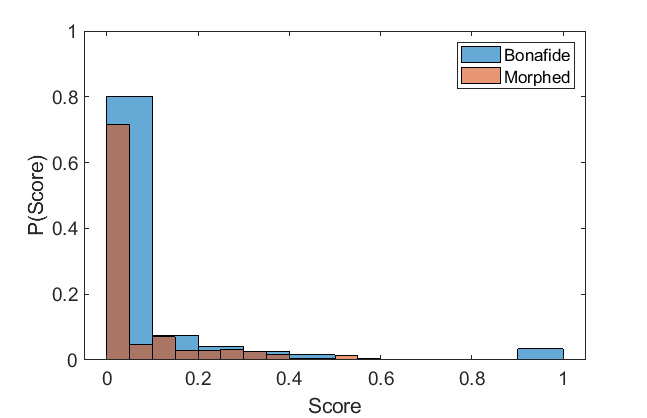}
    \caption{The COTS face comparator scores corresponding to bonafide and morphed test images belonging to the MorGAN dataset (Experiment 2).}
    \label{fig:MORLIMIT}
    
\end{figure}

\begin{figure*}[h]
    \centering
    \includegraphics[width=0.9\textwidth]{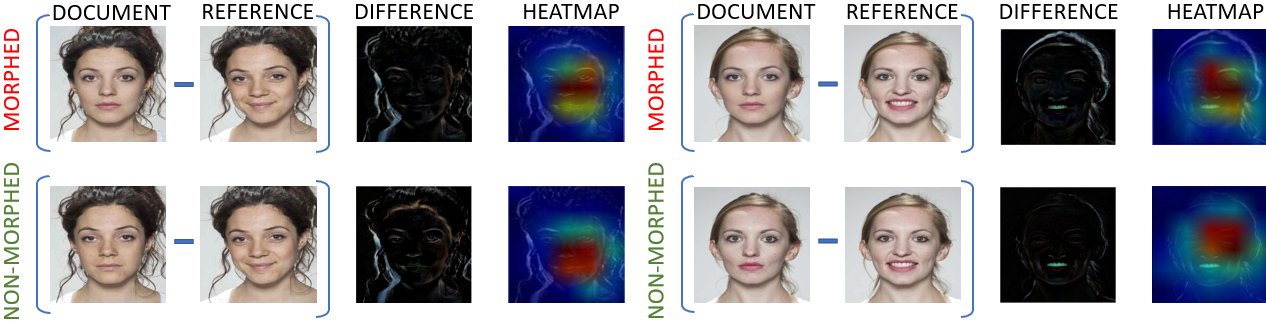}
    \caption{Heatmap visualization of the ``difference images'' for morphed and non-morphed inputs corresponding to two subjects. Note that the difference images contain some discernible information that can be harnessed for recovering some unique information about the second constituent in the morphed image (Experiment 3).}
    \label{fig:Expt3-B}
    
\end{figure*}

  \textbf{Results from Experiment 2:} Table~\ref{Tab:comparison} compares the performance of the proposed method with the two baselines~\cite{DMAD6, DMAD4}. The proposed method achieves BPCER=4.6\% @ APCER=10\%, which outperforms existing methods (\cite{DMAD6}: BPCER=18.5\% @ APCER=10\% and \cite{DMAD4}: BPCER=24.5\% @ APCER=10\%) by a considerable margin when trained on AMSL but tested on MorGAN dataset. However, training on the MorGAN dataset results in poor performance when tested on AMSL dataset. We investigated this issue further and observed an interesting phenomenon. We computed the Pearson correlation coefficient between the document and reference images in the AMSL test set, and when averaged across all the images, it resulted in a mean and standard deviation of $0.91 \pm 0.03$, while for the MorGAN test set, the mean and standard deviation were $0.48 \pm 0.24$, which is almost half of that on the AMSL dataset. We also computed the biometric utility of the MorGAN test set images using the COTS comparator and observed that the biometric similarity scores are low (see Figure~\ref{fig:MORLIMIT}) for both bonafide and morphed images. We suspect the the low degree of correlation between the document and reference images, and the overall low biometric matching utility of the images in the MorGAN dataset, may be responsible for the lower performance of the proposed method in this case. Our method requires adequately high degree of correlation between the document and reference images for successful identity disentanglement.

\begin{figure}[h]
    \centering
    \includegraphics[width=0.485\textwidth]{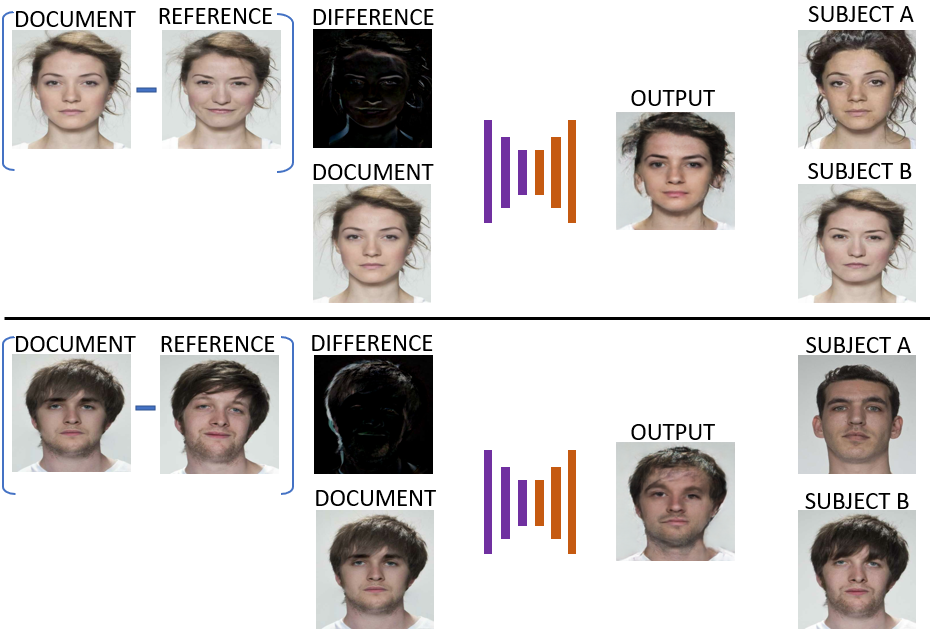}
    \caption{Illustration of outputs generated using the proposed method for decoding information about the ``other'' subject used in morph generation (Experiment 3). Note in both cases the output is visually more similar to Subject A (second subject) compared to Subject B (anchor subject present in reference image).}
    \label{fig:Expt3}
    
\end{figure}

 \textbf{Results from Experiment 3:} In order to recover some information about the second subject who contributed to the morphed image, we utilized the pixel-wise difference images. However, we first needed to ascertain whether the difference images contained any useful information at all. Therefore, we used a similarity visualization technique~\cite{VisSim} to visualize whether the difference image retained any useful information. Figure~\ref{fig:Expt3-B} depicts the visualization in terms of heatmaps (saliency maps) which shows that the difference images contain discriminative information which can be harnessed for decoding the second subject. Figure~\ref{fig:Expt3} illustrates examples of images generated using the proposed method that \textit{qualitatively} appear similar to the second subject (Subject A) compared to the anchor subject (Subject B); the anchor subject pertains to the identity in the reference image. Further, we also analyzed the similarity between the generated outputs and the constituent subjects \textit{quantitatively} using a COTS face comparator. The biometric similarity between the generated outputs and the second subject, as assessed using the true match rate at a false match rate of 1\%, was 23.4\% higher than that between the generated outputs and the first (anchor) subject. Therefore, we note that the proposed method displays tremendous promise for decoding the second identity.  

\section{Summary and Future Work}
\label{Sec:Sum}
In this paper, we proposed a novel differential face morph attack detection framework that disentangles subject identities from a morphed document image conditioned on the trusted reference image. In contrast to existing classification-based approaches, the proposed method formulates the differential MAD problem using an information theoretic framework. We used a conditional generative network to produce an output image from the input (bonafide or morphed) document image. Next, we compared the output image with the reference image using a biometric comparator. The ensuing score is then used for detecting morphs. The method demonstrates a promising first step across different kinds of morph attacks without requiring any supervision about the type of morphing. We achieved best performing results of 3\% BPCER @10\% APCER on intra-dataset evaluation and 4.6\% BPCER @ 10\% APCER on cross-dataset evaluation (Table~\ref{Tab:comparison}). We observed that the proposed method needs well-correlated document and reference image pairs acquired in controlled settings for its robust operation (Section~\ref{Sec:Results}). This is a reasonable requirement given that the success of face morphing is enhanced under these conditions. Furthermore, we used the proposed method to recover some characteristics about the second subject (different from the first subject whose reference image was used). Future work will involve improving the task of de-morphing as well as studying the applicability of the proposed technique to other modalities~\cite{IrisMorph}.

\balance
{\small
\bibliographystyle{ieee}
\bibliography{egbib}

\begin{thebibliography}{10}\itemsep=-1pt

\bibitem{AMSL2}
{AMSL Face Morph Image Data Set}.
\newblock \url{https://omen.cs.uni-magdeburg.de/disclaimer/index.php}.
\newblock [Online accessed: 15th February, 2021].

\bibitem{FaceMorpher}
{Face Morpher}.
\newblock \url{http://www.facemorpher.com/}.
\newblock [Online accessed: 1st July, 2021].

\bibitem{iMARS}
{image Manipulation Attack Resolving Solutions (iMARS)}.
\newblock \url{https://cordis.europa.eu/project/id/883356}.
\newblock [Online accessed: 2nd April, 2021].

\bibitem{Survey1}
K.~B.~Raja et~al.
\newblock {Morphing Attack Detection - Database, Evaluation Platform and
  Benchmarking}.
\newblock {\em IEEE Transactions on Information Forensics and Security}, 2020.

\bibitem{FocusedLRP}
P.~E. Clemens~Seibold, Anna~Hilsmann.
\newblock {Focused LRP: Explainable AI for Face Morphing Attack Detection}.
\newblock {\em IEEE Winter Conference on Applications of Computer Vision
  Workshops}, pages 88--96, 2021.

\bibitem{DMAD1}
N.~Damer, V.~Boller, Y.~Wainakh, F.~Boutros, P.~Terh{\"o}rst, A.~Braun, and
  A.~Kuijper.
\newblock {Detecting Face Morphing Attacks by Analyzing the Directed Distances
  of Facial Landmarks Shifts}.
\newblock {\em {German Conference on Pattern Recognition}}, 2018.

\bibitem{CIEMORGAN}
N.~Damer, F.~Boutros, A.~M. Saladie, F.~Kirchbuchner, and A.~Kuijper.
\newblock {Realistic Dreams: Cascaded Enhancement of GAN-generated Images with
  an Example in Face Morphing Attacks}.
\newblock {\em IEEE 10th International Conference on Biometrics Theory,
  Applications and Systems}, pages 1--10, 2019.

\bibitem{MORGAN}
N.~Damer, A.~M. Saladie, A.~Braun, and A.~Kuijper.
\newblock {MorGAN: Recognition Vulnerability and Attack Detectability of Face
  Morphing Attacks Created by Generative Adversarial Network}.
\newblock {\em IEEE 9th International Conference on Biometrics Theory,
  Applications and Systems}, pages 1--10, 2018.

\bibitem{Magic}
M.~Ferrara, A.~Franco, and D.~Maltoni.
\newblock {The magic passport}.
\newblock {\em IEEE International Joint Conference on Biometrics}, pages 1--7,
  2014.

\bibitem{DMAD3}
M.~Ferrara, A.~Franco, and D.~Maltoni.
\newblock {Face Demorphing}.
\newblock {\em IEEE Transactions on Information Forensics and Security},
  13:1008--1017, 04 2018.

\bibitem{PIX2PIX}
P.~Isola, J.-Y. Zhu, T.~Zhou, and A.~A. Efros.
\newblock {Image-to-Image Translation with Conditional Adversarial Networks}.
\newblock {\em IEEE Conference on Computer Vision and Pattern Recognition},
  pages 5967--5976, 2017.

\bibitem{LMA1}
A.~Makrushin, T.~Neubert, and J.~Dittmann.
\newblock {Automatic Generation and Detection of Visually Faultless Facial
  Morphs}.
\newblock In {\em VISIGRAPP}, 2017.

\bibitem{cgAN}
M.~Mirza and S.~Osindero.
\newblock {Conditional Generative Adversarial Nets}.
\newblock {\em ArXiv}, abs/1411.1784, 2014.

\bibitem{Germanmorphing}
M.~Monroy.
\newblock Laws against morphing.
\newblock \url{https://digit.site36.net/2020/01/10/laws-against-morphing/}.
\newblock {Appeared in Security Architectures and the Police Collaboration in
  the EU 10/01/2020} [Online accessed: 1st April, 2021].

\bibitem{AMSL1}
T.~Neubert, A.~Makrushin, M.~Hildebrandt, C.~Kr{\"a}tzer, and J.~Dittmann.
\newblock {Extended StirTrace benchmarking of biometric and forensic qualities
  of morphed face images}.
\newblock {\em IET Biometrics}, 7:325--332, 2018.

\bibitem{NIST}
M.~Ngan, P.~Grother, K.~Hanaoka, and J.~Kuo.
\newblock {Face Recognition Vendor Test (FRVT) Part 4: MORPH - Performance of
  Automated Face Morph Detection}.
\newblock {\em NISTIR 8292 Draft Supplement}, April 16, 2021.

\bibitem{Asem}
A.~A. Othman and A.~Ross.
\newblock {Privacy of Facial Soft Biometrics: Suppressing Gender But Retaining
  Identity}.
\newblock In {\em {European Conference on Computer Vision Workshops}}, 2014.

\bibitem{FDGAN}
F.~Peng, L.~B. Zhang, and M.~Long.
\newblock {FD-GAN: Face De-Morphing Generative Adversarial Network for
  Restoring Accomplice’s Facial Image}.
\newblock {\em {IEEE Access}}, 7:75122--75131, 2019.

\bibitem{DMAD2}
U.~Scherhag, D.~Budhrani, M.~Gomez-Barrero, and C.~Busch.
\newblock {Detecting Morphed Face Images Using Facial Landmarks}.
\newblock In {\em International Conference on Image and Signal Processing},
  2018.

\bibitem{DMAD_BSIF}
U.~Scherhag, C.~Rathgeb, and C.~Busch.
\newblock {Towards Detection of Morphed Face Images in Electronic Travel
  Documents}.
\newblock In {\em {IAPR 13th International Workshop on Document Analysis
  Systems}}, pages 187--192, 2018.

\bibitem{DMAD4}
U.~Scherhag, C.~Rathgeb, J.~Merkle, and C.~Busch.
\newblock {Deep Face Representations for Differential Morphing Attack
  Detection}.
\newblock {\em IEEE Transactions on Information Forensics and Security},
  15:3625--3639, 2020.

\bibitem{LMA2}
C.~Seibold, W.~Samek, A.~Hilsmann, and P.~Eisert.
\newblock Detection of face morphing attacks by deep learning.
\newblock In C.~Kraetzer, Y.-Q. Shi, J.~Dittmann, and H.~J. Kim, editors, {\em
  Digital Forensics and Watermarking}, pages 107--120, 2017.

\bibitem{IrisMorph}
R.~Sharma and A.~Ross.
\newblock {Image-level Iris Morph Attack}.
\newblock {\em {IEEE 28th International Conference on Image Processing}}, 2021.

\bibitem{DMAD6}
S.~Soleymani, A.~Dabouei, F.~Taherkhani, J.~Dawson, and N.~M. Nasrabadi.
\newblock {Mutual Information Maximization on Disentangled Representations for
  Differential Morph Detection}.
\newblock {\em IEEE Winter Conference on Applications of Computer Vision},
  pages 1731--1741, 2021.

\bibitem{History}
M.~Steyvers.
\newblock {Morphing techniques for manipulating face images}.
\newblock {\em Behavior Research Methods, Instruments, \& Computersc},
  31:359--369, 06 1999.

\bibitem{VisSim}
A.~{Stylianou}, R.~{Souvenir}, and R.~{Pless}.
\newblock {Visualizing Deep Similarity Networks}.
\newblock In {\em IEEE Winter Conference on Applications of Computer Vision},
  pages 2029--2037, 2019.

\bibitem{Survey2}
S.~Venkatesh, R.~Ramachandra, K.~Raja, and C.~Busch.
\newblock {Face Morphing Attack Generation \& Detection: A Comprehensive
  Survey}.
\newblock {\em IEEE Transactions on Technology and Society}, 2021.

\bibitem{MIP}
H.~Zhang, S.~Venkatesh, R.~Ramachandra, K.~Raja, N.~Damer, and C.~Busch.
\newblock {MIPGAN - Generating Robust and High Quality Morph Attacks Using
  Identity Prior Driven GAN}.
\newblock {\em IEEE Transactions on Biometrics}, 2021.

\end{thebibliography}
}

\end{document}